\documentclass[final]{cvpr}

\usepackage{times}
\usepackage{epsfig}
\usepackage{graphicx}
\usepackage{amsmath}
\usepackage{amssymb}
\usepackage{caption}
\usepackage{subcaption}
\usepackage{ulem}
\usepackage{tabularx}
\usepackage{booktabs}

\usepackage[pagebackref=true,breaklinks=true,colorlinks,bookmarks=false]{hyperref}

\def\cvprPaperID{7130} 
\def\confYear{CVPR 2021}

\newcommand\blfootnote[1]{%
  \begingroup
  \renewcommand\thefootnote{}\footnote{#1}%
  \addtocounter{footnote}{-1}%
  \endgroup
}

\begin{document}

\title{Rethinking FUN: Frequency-Domain Utilization Networks}
\author{
 Kfir Goldberg\textsuperscript{1,2} 
 \qquad Stav Shapiro\textsuperscript{2}
 \qquad Elad Richardson\textsuperscript{2} 
 \qquad Shai Avidan\textsuperscript{1} \\ 
 \textsuperscript{1}Tel-Aviv University \qquad  \textsuperscript{2}Penta-AI
}


\maketitle
\begin{abstract}
The search for efficient neural network architectures has gained much focus in recent years, where modern architectures focus not only on accuracy but also on inference time and model size. Here, we present FUN, a family of novel Frequency-domain Utilization Networks. These networks utilize the inherent efficiency of the frequency-domain by working directly in that domain, represented with the Discrete Cosine Transform. Using modern techniques and building blocks such as compound-scaling and inverted-residual layers we generate a set of such networks allowing one to balance between size, latency and accuracy while outperforming competing RGB-based models. Extensive evaluations verifies that our networks present strong alternatives to previous approaches. Moreover, we show that working in frequency domain allows for dynamic compression of the input at inference time without any explicit change to the architecture.
\end{abstract}

\section{Introduction}
\blfootnote{Code available at\ \ \url{https://github.com/kfir99/FUN}}
The introduction of well-designed convolutional neural networks (CNNs) has revolutionized the computer vision field~\cite{krizhevsky2012imagenet, alom2018history,conneau2016very, szegedy2015going, he2016deep}. While early efforts in the design of such networks focused mostly on maximizing accuracy, recent works \cite{tan2019efficientnet,howard2017mobilenets,radosavovic2020designing,wu2019fbnet,tan2019mnasnet} have shown the importance of architectures that are not only accurate, but also efficient. More specifically, these works focus on lowering the number of floating point operations (FLOPs), inference latency and the number of model parameters while maximizing model accuracy.

\begin{figure}
    \includegraphics[scale=0.4]{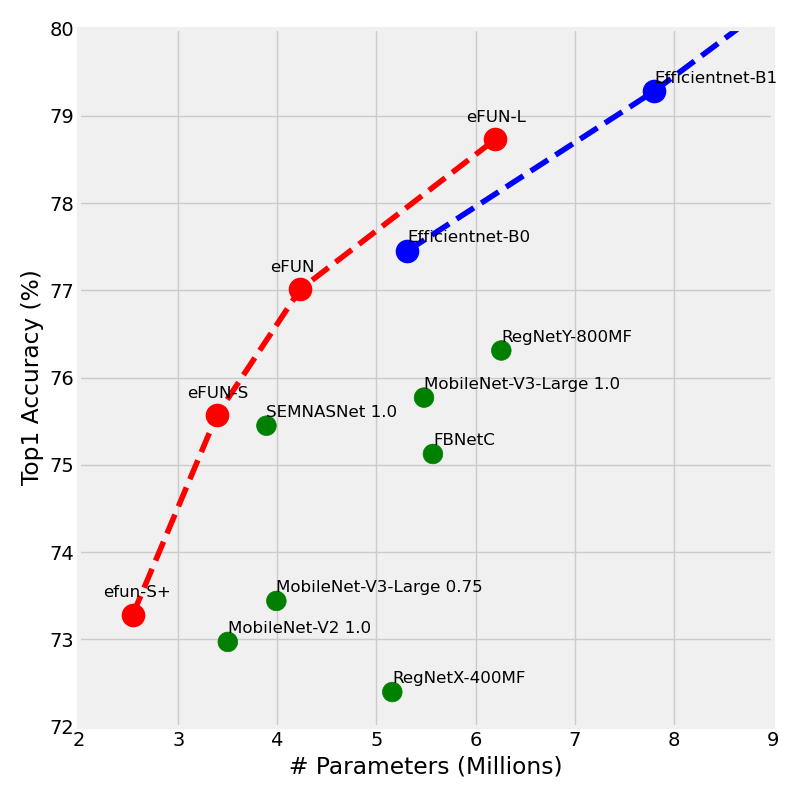}
    \caption{ImageNet Accuracy vs Model Size}
    \label{imagenet_acc_vs_size}
\end{figure}

An increasingly popular avenue for research on CNNs efficiency
is to leverage frequency domain representations of images, most commonly achieved using the Discrete Cosine Transform (DCT). Being part of the JPEG codec, the DCT is both widely used and convenient to work with. As a result, previous works used the DCT coefficients as inputs to CNNs, achieving faster inference ~\cite{gueguen2018faster, Ehrlich_2019_ICCV} and more control ~\cite{xu2020learning} over the input size.
\\
We build upon these works by introducing a family of novel Frequency-domain Utilization Networks (FUN). As the name suggests, FUN networks utilize the advantages inherent in the frequency domain representations. We present a family of architectures that allow for trade-offs between model accuracy, model size, and inference latency, while also allowing for different levels of input compression.\\

Motivated by the results presented in \cite{gueguen2018faster}, we explore similar design principles in the widely used ResNet architectures. The resulting frequency domain architecture, named ResFUN, is superior in terms of efficiency and model size to both the original ResNet and the modified DCT based architecture in \cite{gueguen2018faster}. Next, we apply these design principles to more efficiency oriented class of architectures ~\cite{sandler2018mobilenetv2,tan2019efficientnet} whose main building block is the Inverted Residual Block (MBConv). The resulting DCT-based architecture, named eFUN, is significantly faster and lighter than the equivalent RGB-based architectures, while achieving comparable, and at times, even better accuracy. We further extend our FUN family by using the compound scaling method~\cite{tan2019efficientnet}, generating models with a range of sizes, speeds, and accuracy, providing different degrees of freedom when selecting a model for different computational budgets. Figure~\ref{imagenet_acc_vs_size} summarizes our performance on the ImageNet classification task, showing that eFUN outperforms other competing models.\\

Extensive evaluations are performed to measure the effectiveness and benefits of our eFUN family of models compared to highly-optimized RGB-based networks. For example, on the ImageNet classification task, our base model, eFUN, is $20\%$ smaller and $65\%$ faster than EfficientNet-B0, while dropping only a small percentage of accuracy. Notably our results show that due to the inherent properties of the DCT representation, one can remove more than half of the input channels of a model trained on ImageNet with a drop of less than $1\%$ in top-1 accuracy with no additional training. 

Our main contributions are:
\begin{itemize}
    \item A novel, DCT-based, CNN architecture achieving state-of-the-art results on ImageNet classification.
    \item A family of frequency-based models constructed using the compound scaling method introduced in~\cite{tan2019efficientnet}.
\end{itemize}

\section{Related Work} \label{Related Work}

\paragraph{Efficient CNNs}
In recent years, CNNs have significantly improved, becoming increasingly more accurate over time. However, this accuracy has come at a cost in the form of slower, larger architectures~\cite{touvron2020fixing, xie2020self, huang2019gpipe}.
As these CNNs have gotten bigger, we have begun reaching hardware capacity limitations. For example, GPIPE \cite{huang2019gpipe} with over $560$M parameters requires a dedicated infrastructure for training. Therefore, recent works \cite{tan2019mnasnet, howard2017mobilenets, sandler2018mobilenetv2, tan2019efficientnet, zhang2017shufflenet, i2016squeezenet} have focused on designing more efficient architectures without compensating for accuracy. 
While many works have done so using more standard techniques such as pruning \cite{frankle2018lottery} and quantization \cite{wu2016quantized}, it has become common to design highly-optimized architectures obtained using a Neural Architecture Search (NAS) aimed at optimizing the FLOP count as a proxy for inference latency that is agnostic of the underlying hardware.
Contrary to these works, we approach the efficiency problem by changing the representation of the data, from RGB to DCT coefficients, and design efficient architectures operating on the DCT inputs.

\paragraph{Frequency Domain Based Networks}
The utilization of the frequency domain for CNNs has been studied in the past for various reasons.
\cite{gueguen2018faster} have shown that feeding DCT inputs to ResNet-based CNNs can lead to a significant speedup in inference latency while maintaining accuracy close to that of RGB-based models. The speedup provided by their architectures is due to the shallower network and the lower spatial size of the DCT inputs. 
We extend this work, utilizing more modern architectural building blocks and achieving better efficiency in terms of inference latency, model size, and input size. 
\cite{xu2020learning} trade-off the spatial size of the input with the number of DCT coefficients used to represent each block in the DCT. They suggest using DCT inputs with larger spatial dimensions ($448\times448$ compared to $224\times224$ for RGB), while keeping only a certain number of DCT coefficients from each block. We show that our FUN models can work well even when drastically reducing the number of DCT channels used for each block, without any need to change the spatial size of the input, thus enabling the use of highly compressed inputs.
Finally, \cite{ulicny2020harmonic} propose using a new block, the Harmonic block, in RGB-based architectures, which relies on the DCT filters.
\section{Frequency-Domain Utilization Networks}~\label{FUN framework}
In this section we introduce a set of novel architectures operating on inputs in the frequency domain. We first describe the preprocessing required for the DCT inputs, followed by the network design process for the different architectures.

\begin{figure}[b]
    \resizebox{1.0\columnwidth}{!}{\includegraphics[scale=0.5]{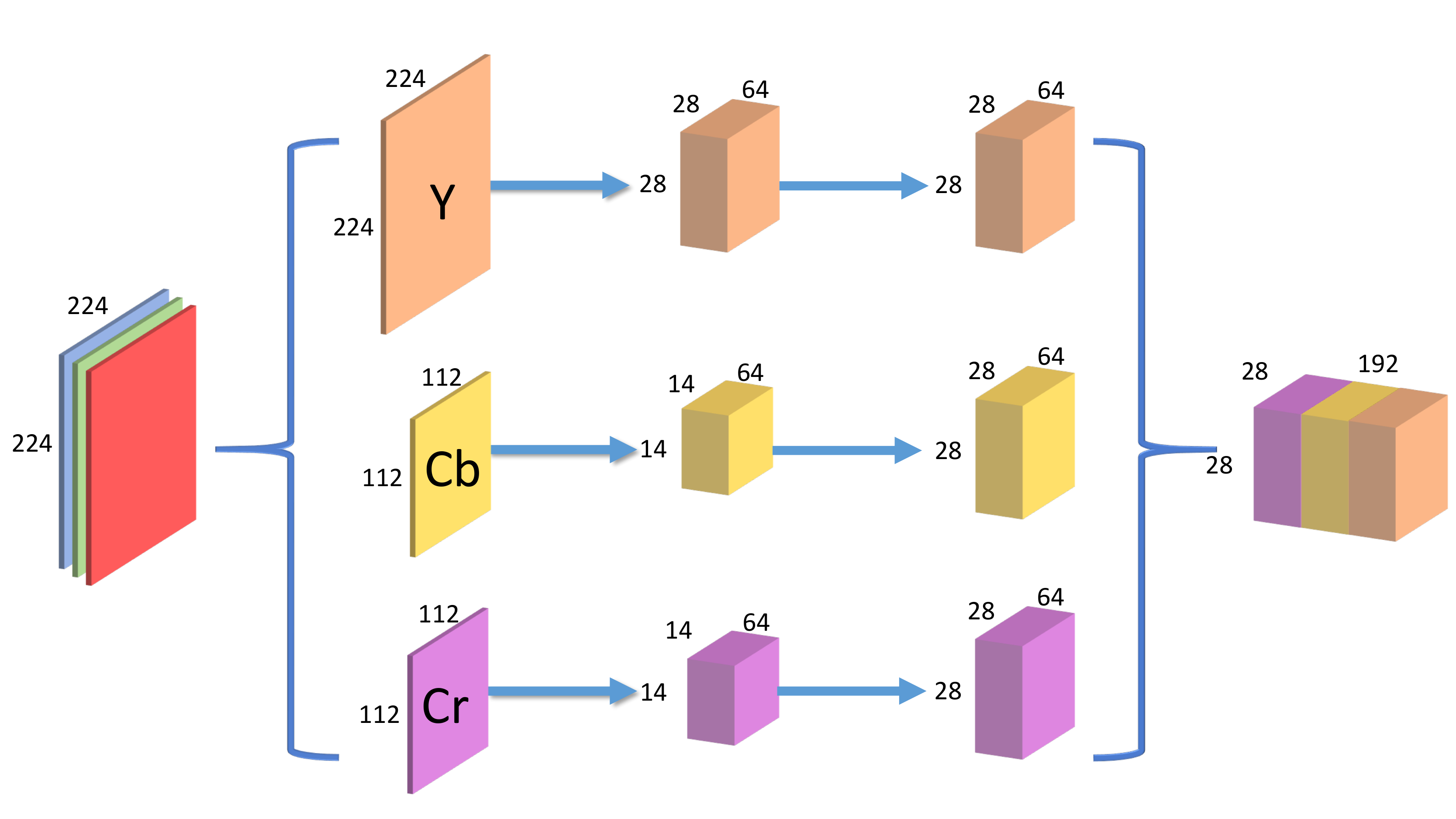}}
    \caption{RGB to DCT preprocessing for FUN}
    \label{fig:fun_preprocess}
    \vspace{-2mm}
\end{figure}

\begin{figure*}
    \centering
    \begin{subfigure}{\textwidth}
        \centering
        \includegraphics[width=0.75\textwidth]{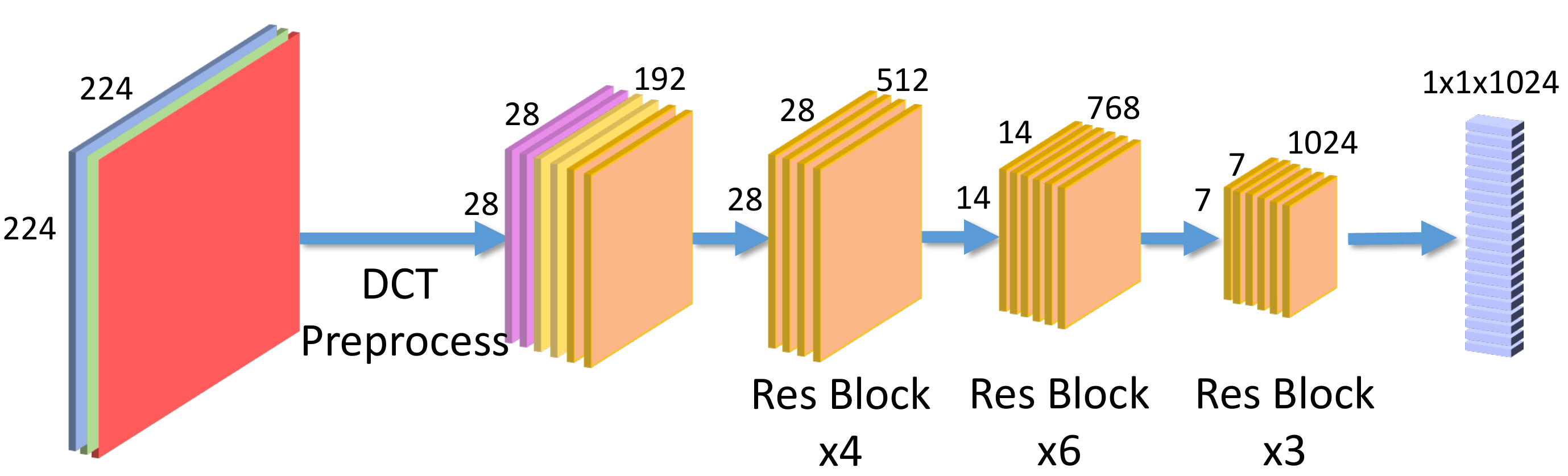}
        \caption{}
        ~\label{fig:resfun_arch}
    \end{subfigure}\\
    \begin{subfigure}{\textwidth}
        \centering
        \includegraphics[width=0.75\textwidth]{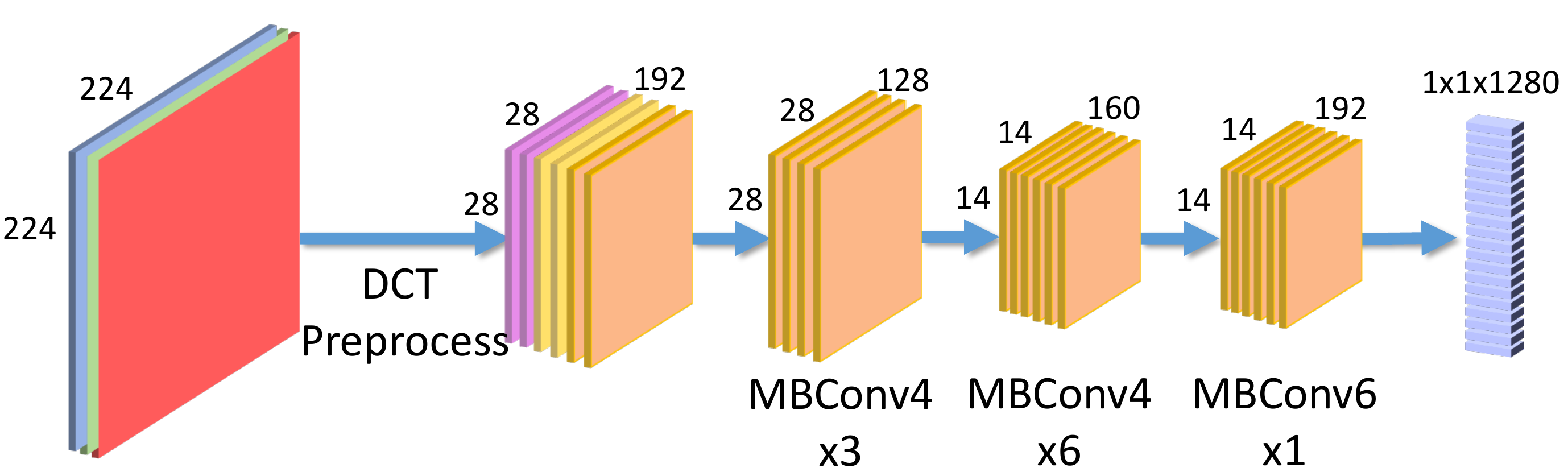}
        \caption{}
        ~\label{fig:efun_arch}
    \end{subfigure}
    \newline
    \caption{FUN architectures, (\subref{fig:resfun_arch}) presents ResFUN, (\subref{fig:efun_arch}) presents eFUN.}
    \label{archs}
    \vspace{-2mm}
\end{figure*}

\subsection{The DCT Preprocessing Pipeline}~\label{preprocessing}
Our preprocessing stage is described in Figure \ref{fig:fun_preprocess}. Similar to \cite{gueguen2018faster}, the RGB-represented input image is first converted to its corresponding YCbCr representation, representing the luminance (Y) and chroma (Cb, Cr) of the image. The two chroma channels are then down-sampled by a factor of $2$, as is done in the JPEG codec. Each of the three channels is then split into $8\times8$ blocks and passed through the DCT converting each to its frequency-domain representation. Next, the two chroma channels are up-sampled to have the same spatial size as the luminance channel. Finally, the three channels are stacked over the frequency domain.

\subsection{The ResFUN Architecture}
First, as an illustrative example, we present the ResFUN architecture described in Figure \ref{fig:resfun_arch}. Similar to the standard ResNet architecture \cite{he2016deep}, ResFUN is constructed using Residual Blocks. In the case of ResFUN, however, the input layer must be changed to accommodate the DCT-represented inputs. In a similar fashion to \cite{gueguen2018faster}, we remove the first convolutional layer and first Residual Block of the ResNet, corresponding to the initial $\times8$ down-sampling of the original RGB input image, to fit the $28\times28$ spatial size of the DCT-based inputs.

Additionally, to further reduce the number of parameters in our model, we decrease the number of filters in each of the ResNet stages --- from $512$, $1024$, $2048$ in ResNet-50 to $512$, $768$, $1024$ in our ResFUN architecture, respectively. Doing so reduces the model size by a factor of $2.5$, from $25.5$M parameters in ResNet-50, to $10.4$M in our ResFUN model. This additional reduction in model size is possible since our DCT inputs are already a processed, more compact representation than the RGB inputs, for which the standard ResNets are designed.

\subsection{The eFUN Architecture}
Having successfully converted the ResNet architecture to operate on DCT-represented inputs, we then turn to more efficiency-oriented architectures. Specifically, we focus on the class of architectures utilizing the Inverted Residual Block (MBConv)~\cite{sandler2018mobilenetv2,howard2019searching, tan2019mnasnet, tan2019efficientnet, radosavovic2020designing}. A notable example of such architectures is the EfficientNet~\cite{tan2019efficientnet} family of architectures, which additionally introduces the compound model scaling technique. There, one can generate scaled-up versions of a baseline model by intelligently balancing the width, depth, and input resolution to maximize performance for a given computational budget.
Following these recent architectural advancements, we present a family of FUN models, efficientFUN (eFUN). Our baseline eFUN model, illustrated in Figure~\ref{fig:efun_arch} is inspired by EfficientNet-B0, but with several key differences. 

\paragraph{Shallower Architecture} As EfficientNet-B0 receives RGB inputs of size $224\times224\times3$, their input is only down-sampled to a spatial dimension of $28\times28$ after the first six layers. Conversely, our eFUN models are given DCT inputs already with a spatial resolution of $28\times28$.
The smaller input resolution enables us to create a shallower network, leading to lower inference latency compared to the larger EfficientNet counterpart.  

\paragraph{Wider Bottleneck} In EfficientNet-B0, the stages with equivalent input resolutions to the DCT domain are $5$ and $6$. To compensate for the loss of representation strength from stages $1$ through $4$, we widen stages $5$ and $6$ of EfficientNet-B0 channels from $80$ and $112$, to $128$ and $160$ respectively.

\paragraph{Scaled-Down Models} Given the base model constructed above, the compound scaling method is leveraged to generate additional architectures. In contrast to ~\cite{tan2019efficientnet} however, we scale \textit{down} our eFUN model and generate the smaller eFUN-S+ and eFUN-S models in addition to scaling up to obtain the larger eFUN-L model. 

\subsection{The Benefits of FUN}

The shallower design of the eFUN and ResFUN architectures is a direct consequence of working in the frequency domain. Specifically in the case of DCT, this results in a wider but shallower network. In the following sections we show that this design change in the eFUN architecture is superior to the original design in both accuracy and model size. Our results are in agreement with  \cite{zagoruyko2016wide} where it was shown that the wider but shallower deep residual networks outperformed previous deeper but thinner versions of the same networks.

Additionally, due to the parallel compute paradigm in modern GPUs, a known rule of thumb is that, all else being equal, shallow and wide beats deep and narrow in terms of inference latency. For our eFUN network, this can lead to between $30\%$ and $80\%$ inference time improvements compared to baseline RGB architectures.

Finally, FUNs allow for dynamic reduction of the input size. Unlike the RGB or YCbCr representations, where all 3 input channels have similar importance, in DCT coefficients the lower frequencies tend to have a larger visual importance. This insight is leveraged to show how the eFUN architectures can work well even when significantly reducing the number of input channels of a trained model, with no additional training or modifications needed.

\subsection{Implementation Details}\label{subsec:imp}
All of our FUN models take as input 224x224 RGB images, which first pass through the preprocessing presented in Section~\ref{preprocessing}. We use the computed DCT of each input image, with the same configuration as the standard JPEG DCT, and use JPEG image quality set to 100, meaning the DCT coefficients are not further compressed. The DCT coefficients obtained from the JPEG pipeline are in the shape of 28x28x64 for the Y channel, and 14x14x64 for each of the Cb and Cr channels. The chroma channels are then up-sampled and all three channels are stacked over the frequency dimension, resulting in a final input of size 28x28x192.

The eFUN models are trained on ImageNet using a similar settings to that of~\cite{tan2019efficientnet}:
The RMSProp optimizer is used with decay of 0.9 and momentum of 0.9; weight decay of 1e-5; and a learning rate scheduler with an initial learning rate of 0.048 with a decay of 0.97 every 2.4 epochs.
We also use stochastic depth with a drop probability of 0.2 for eFUN-S+, eFUN-S and eFUN, and 0.3 drop probability for eFUN-L. The models are trained for 450 epochs using 4 NVIDIA V100 GPUs and a batch size of 512 and inference latency of all models is measured using a single NVIDIA V100 GPU with a batch size of 1. All latency measurements are computed as if the images were already loaded in the relevant representation (RGB/DCT), thus the timing improvement possible due to reduced JPEG decompression time for DCT representation is not taken into account.
The results for RGB-based models are reported using the same implementation \footnote{\url{https://github.com/rwightman/pytorch-image-models}} where EfficientNet-B0 and EfficientNet-B1 are trained using the same settings as the eFUN models.
Additionally, Our ResFUN model is trained using the SGD optimizer with a learning rate of 0.1 for 90 epochs, 0.01 for 20 epochs, and 0.001 for 20 epochs.

\begin{table}
  \vskip 0.1in
    \centering                                                                  
    \begin{tabular}{l c c c}
    \toprule 
    Model & Top-1 Acc. & \# Parameters (M) & Images / sec.\\
    \midrule
    \bf ResFUN & 
    \multicolumn{1}{c}{$72.8\%$} &
    \multicolumn{1}{c}{$10.4$} &
    \multicolumn{1}{c}{$167$} \\
    ResNet18 & 
    \multicolumn{1}{c}{$69.8\%$} &
    \multicolumn{1}{c}{$11.7$} &
    \multicolumn{1}{c}{$330$} \\
    ResNet34 & 
    \multicolumn{1}{c}{$73.3\%$} &
    \multicolumn{1}{c}{$21.8$} &
    \multicolumn{1}{c}{$180$} \\
    \bottomrule 
    \end{tabular}   
    \caption{ResFUN performance compared to ResNets with different depths}
    \label{tab:resnet_results}     
\end{table}


\section{Experiments} \label{Experiments}
In this section an evaluation of the FUN models introduced in Section \ref{FUN framework} is performed.
We start by evaluating our ResFUN architecture in comparison to the standard ResNet architectures.
We then conduct an extensive evaluation of our eFUN family of models, and compare them to a wide variety of efficiency-oriented RGB-based models.
We continue by investigating the transfer-learning capabilities of our models compared to those of RGB-based models.
Finally, we perform an ablation study to explore different possible architectural choices.

\begin{table*}
  \vskip 0.1in
    \centering                                                                      
    \begin{tabular}{l c c c c}
    \toprule 
    Model & Top-1 Acc. & \# Parameters (M) & Images / sec. & \#FLOPs (M)\\
    \midrule
    \bf eFUN-S+ & 
    \multicolumn{1}{c}{$73.3\%$} &
    \multicolumn{1}{c}{$2.5$} &
    \multicolumn{1}{c}{$145$} &
    \multicolumn{1}{c}{$520$} \\
    RegNetX-400MF \cite{radosavovic2020designing} & 
    \multicolumn{1}{c}{$72.4\%$} &
    \multicolumn{1}{c}{$5.2$} &
    \multicolumn{1}{c}{$70$} &
    \multicolumn{1}{c}{$400$} \\
    MobileNet-V2 1.0 \cite{sandler2018mobilenetv2} & 
    \multicolumn{1}{c}{$73\%$} &
    \multicolumn{1}{c}{$3.5$} &
    \multicolumn{1}{c}{$148$} &
    \multicolumn{1}{c}{$300$} \\
    MobileNet-V3-Large 0.75 \cite{howard2019searching} & 
    \multicolumn{1}{c}{$73.4\%$} &
    \multicolumn{1}{c}{$4$} &
    \multicolumn{1}{c}{$95$} &
    \multicolumn{1}{c}{$150$} \\
    \hline
    \bf eFUN-S & 
    \multicolumn{1}{c}{$75.6\%$} &
    \multicolumn{1}{c}{$3.4$} &
    \multicolumn{1}{c}{$132$} &
    \multicolumn{1}{c}{$600$}\\
    FBNet-C\cite{wu2019fbnet}&
    \multicolumn{1}{c}{$75.1\%$} &
    \multicolumn{1}{c}{$5.6$} &
    \multicolumn{1}{c}{$122$} &
    \multicolumn{1}{c}{$375$} \\
    SEMNASNet 1.0 \cite{tan2019mnasnet} & 
    \multicolumn{1}{c}{$75.4\%$} &
    \multicolumn{1}{c}{$3.9$} &
    \multicolumn{1}{c}{$122$} &
    \multicolumn{1}{c}{$300$} \\
    MobileNet-V3-Large 1.0 \cite{howard2019searching} &
    \multicolumn{1}{c}{$75.8\%$} &
    \multicolumn{1}{c}{$5.5$} &
    \multicolumn{1}{c}{$94$} & 
    \multicolumn{1}{c}{$220$} \\
    RegNetY-800MF \cite{radosavovic2020designing}& 
    \multicolumn{1}{c}{$76.3\%$} &
    \multicolumn{1}{c}{$6.3$} &
    \multicolumn{1}{c}{$77$} &
    \multicolumn{1}{c}{$800$} \\
    \hline
    \bf eFUN &
    \multicolumn{1}{c}{$77\%$} &
    \multicolumn{1}{c}{$4.2$} &
    \multicolumn{1}{c}{$124$} & 
    \multicolumn{1}{c}{$850$} \\
    EfficientNet-B0 \cite{tan2019efficientnet}& 
    \multicolumn{1}{c}{$77.4\%$} &
    \multicolumn{1}{c}{$5.3$} &
    \multicolumn{1}{c}{$77$} &
    \multicolumn{1}{c}{$400$} \\ 
    \hline
    \bf eFUN-L & 
    \multicolumn{1}{c}{$78.8\%$} &
    \multicolumn{1}{c}{$6.2$} &
    \multicolumn{1}{c}{$101$}  &
    \multicolumn{1}{c}{$1,600$} \\
    EfficientNet-B1 \cite{tan2019efficientnet} & 
    \multicolumn{1}{c}{$79.3\%$} &
    \multicolumn{1}{c}{$7.8$} &
    \multicolumn{1}{c}{$56$} &
    \multicolumn{1}{c}{$700$} \\
    \bottomrule 
    \end{tabular}   
    \caption{eFUN performance on ImageNet. CNNs with similar top-1 accuracy are grouped together for clear comparison.}
    \label{tab:imagenet_results}     
\end{table*}

\subsection{ImageNet Results on ResFUN}
We first show results obtained on the ImageNet classification task using our constructed ResFUN architecture compared to standard ResNet architectures operating on inputs represented in RGB.
Results are summarized in Table~\ref{tab:resnet_results}. While the RGB-based ResNets offer different trade-offs between accuracy and model size by changing the depth of the model, our frequency-based ResFUN offers appealing advantages. Compared to the smallest ResNet model, ResNet-18, ResFUN is both shallower and uses less filters in each block while achieving a higher top-1 accuracy by approximately $3\%$ and being smaller in size. Although the larger ResNet-34 architecture out-performs our model with respect to top-1 accuracy, the ResFUN architecture provides an attractive alternative being $\times2$ smaller than the standard ResNet-34 architecture. These comparisons show that the ResFUN architecture manages to be smaller than the ResNet18 while being almost as accurate as the ResNet34, making ResFUN significantly more efficient than the ResNet architectures.

\subsection{ImageNet Results on eFUN}~\label{imagenet results section}
\begin{figure}
    \includegraphics[scale=0.4]{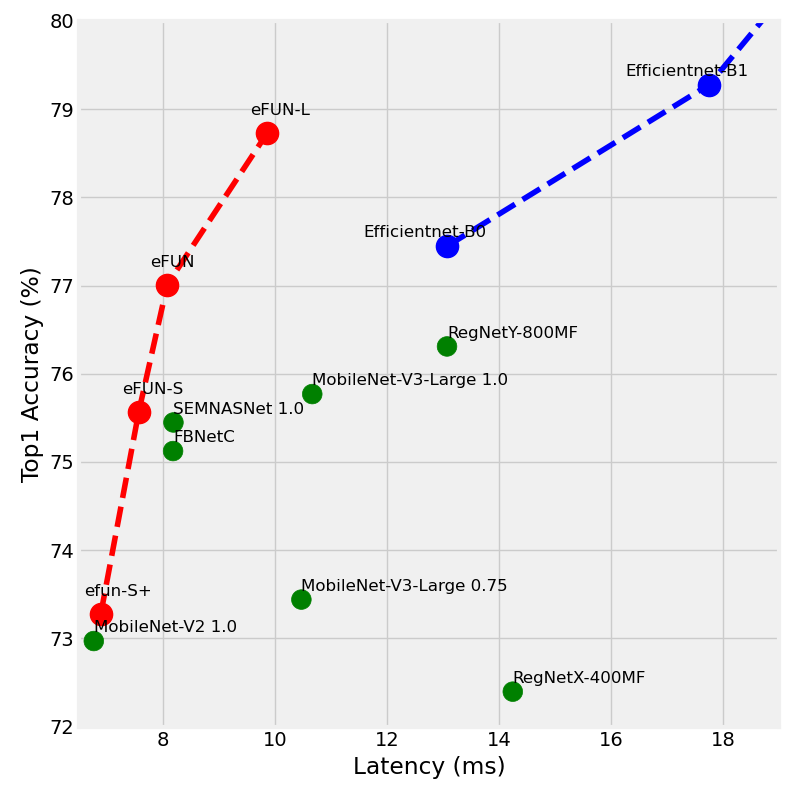}
    \caption{ImageNet Accuracy vs. Latency.}
    \label{imagenet_acc_vs_latency}
\end{figure}

The results of our eFUN models are presented in Table~\ref{tab:imagenet_results} and show that the eFUN model is $20\%$ smaller and $65\%$ faster than the highly-efficient, state-of-the-art EfficientNet-B0 model, while being very close in terms of accuracy ($77.4\%$ for EfficientNet-B0 compared to $77\%$ eFUN).
Building on the eFUN model, we use EfficientNet's compound scaling method, allowing us to scale up and scale down the eFUN architecture to provide flexibility with respect to the desired computational budget. In particular, the smallest architecture generated by compound scaling of the eFUN model is the eFUN-S+ architecture, which is $30\%$ smaller and $0.3\%$ more accurate than the comparable MobileNet-V2-100 model. Although both models obtain similar inference latency (Figure~\ref{imagenet_acc_vs_latency}), MobileNet-V2 was specifically designed and optimized for inference latency, while eFUN-S+ is inherently fast due to the advantages offered by FUN, while also being smaller and more accurate than the former.

Table~\ref{tab:imagenet_results} shows that while  eFUN models are typically faster than similar RGB-based models at inference time, their FLOP count is significantly higher. FLOPs are usually used as a proxy to estimate the run time of models on different CPUs without measuring on a specific setup. Our results show that FLOPs are not a good indicator for inference latency when measured on GPUs, due to their highly paralleled nature. Since eFUN models are generally significantly shallower ($10-13$ layers) than EfficientNet models ($18$ layers for EfficientNet-B0), their execution on GPUs, proportional to the depth of the network, is faster.

\begin{table}   
  \vskip 0.1in
    \centering
    \resizebox{1.0\columnwidth}{!}{ 
    \begin{tabular}{l c c c c}
    \toprule 
    Dataset & Train size & Test Size & \# Classes \\
    \midrule
    CIFAR-100 \cite{krizhevsky2009learning} &
    \multicolumn{1}{c}{50,000} &
    \multicolumn{1}{c}{10,000} &
    \multicolumn{1}{c}{100} \\
    Stanford Cars \cite{krause2013collecting} &
    \multicolumn{1}{c}{8,144} &
    \multicolumn{1}{c}{8,041} &
    \multicolumn{1}{c}{196} \\
    FGVC Aircraft \cite{maji2013fine} &
    \multicolumn{1}{c}{6,667} &
    \multicolumn{1}{c}{3,333} &
    \multicolumn{1}{c}{100} \\
    \bottomrule 
    \end{tabular}
    }
    \caption{Transfer Learning Datasets.}
    \label{tab:transfer_learning_datasets}
\end{table}

\begin{figure*}
\setlength{\tabcolsep}{1pt}
\centering
    \begin{tabular}{c c c}
        \includegraphics[width=0.33\textwidth]{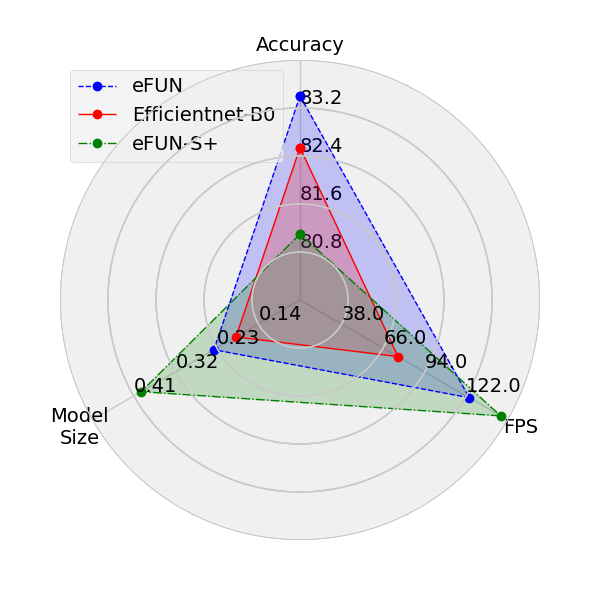}&
        \includegraphics[width=0.33\textwidth]{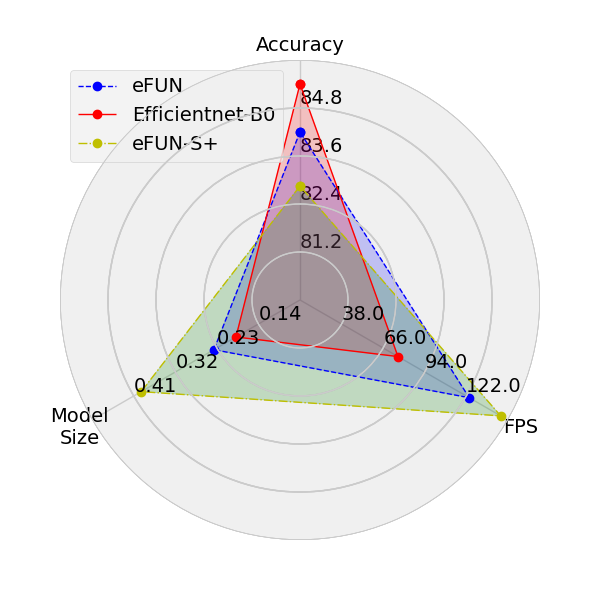}&        
        \includegraphics[width=0.33\textwidth]{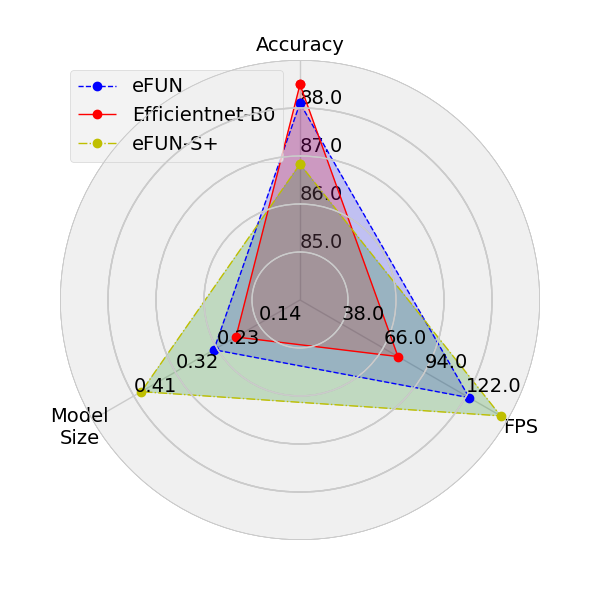}
    \end{tabular}
    \caption{Transfer learning results, from left to right: CIFAR100~\cite{krizhevsky2009learning}, FGVC aircraft~\cite{maji2013fine}, Stanford Cars~\cite{krause2013collecting}.}
    \label{fig:transfer_learning}
\end{figure*}

\subsection{Transfer Learning Results for FUN Architectures}
To show that FUN generalizes for different tasks, we present an evaluation of our models on three commonly used transfer learning datasets, which are presented in Table \ref{tab:transfer_learning_datasets}.
For these experiments, the same settings defined in Section~\ref{subsec:imp} are applied, where each model is first pretrained on ImageNet and then fine-tuned on its task.

Figure \ref{fig:transfer_learning} illustrates the trade-off between model accuracy, model size and inference latency for three transfer learning datasets. For each model, we report results for top-1 accuracy, inference speed (in images per second), and model size (in number of parameters). One can see that eFUN models significantly outperform the competing models in terms of model size and inference latency, while providing comparable accuracy.
Specifically, when considering the popular CIFAR-100 dataset~\cite{krizhevsky2009learning}, the eFUN model outperforms the EfficientNet-B0 model by $0.9\%$ accuracy, while still being $20\% $smaller and $65\%$ faster. Our eFUN-S+ model is lower in accuracy compared to EfficientNet-B0 model ($82.5\%$ for EfficientNet-B0 compared to $81.1\%$ for eFUN-S+), but is $53\%$ smaller and $91\%$ faster.


\subsection{Architecture Ablation Study}

\begin{table}
    \setlength{\tabcolsep}{2.2pt}
    \centering
    \resizebox{1.0\columnwidth}{!}{ 
    \begin{tabular}{c c c c c}
    \toprule
    Stage & 
    Operator &
    Resolution & 
    \# Channels  &
    \# Layers\\
    $i$ &
    $\mathcal{\hat{F}}_i$ &
    $\mathcal{\hat{H}}_i\times\mathcal{\hat{W}}_i$ &
    $\mathcal{\hat{C}}_i$ &
    $\mathcal{\hat{L}}_i$ \\
    \midrule
    $1$ & 
    \multicolumn{1}{c}{MBConv1, k3x3} &
    \multicolumn{1}{c}{$28\times28$} &
    \multicolumn{1}{c}{16} &
    \multicolumn{1}{c}{1}  \\ 
    $2$ & 
    \multicolumn{1}{c}{MBConv6, k3x3} &
    \multicolumn{1}{c}{$28\times28$} &
    \multicolumn{1}{c}{24} &
    \multicolumn{1}{c}{2}  \\ 
    $3$ & 
    \multicolumn{1}{c}{MBConv6, k5x5} &
    \multicolumn{1}{c}{$28\times28$} &
    \multicolumn{1}{c}{40} &
    \multicolumn{1}{c}{2}  \\ 
    $4$ & 
    \multicolumn{1}{c}{MBConv6, k3x3} &
    \multicolumn{1}{c}{$28\times28$} &
    \multicolumn{1}{c}{80} &
    \multicolumn{1}{c}{3}  \\
    $5$ & 
    \multicolumn{1}{c}{MBConv6, k5x5} &
    \multicolumn{1}{c}{$14\times14$} &
    \multicolumn{1}{c}{112} &
    \multicolumn{1}{c}{3}  \\ 
    $6$ & 
    \multicolumn{1}{c}{MBConv6, k5x5} &
    \multicolumn{1}{c}{$14\times14$} &
    \multicolumn{1}{c}{192} &
    \multicolumn{1}{c}{4}  \\ 
    $7$ & 
    \multicolumn{1}{c}{MBConv6, k3x3} &
    \multicolumn{1}{c}{$7\times7$} &
    \multicolumn{1}{c}{320} &
    \multicolumn{1}{c}{1}  \\ 
    $8$ & 
    \multicolumn{1}{c}{Conv1x1 \& Pooling \& FC} &
    \multicolumn{1}{c}{$7\times7$} &
    \multicolumn{1}{c}{1280} &
    \multicolumn{1}{c}{1} \\
    \bottomrule
    \end{tabular}
    }
\caption{The eFUN-variant-A architecture. Each row describes a stage, $i$, with $\hat L_i$ layers, input resolution $\langle \hat H_i, \hat W_i \rangle$ and output channels $\hat C_i$.}
\label{tb:ablation_effnet_b0}
\end{table}

\begin{table}
    \setlength{\tabcolsep}{2.2pt}
    \centering
    \resizebox{1.0\columnwidth}{!}{ 
    \begin{tabular}{c c c c c}
    \toprule
    Stage & 
    Operator &
    Resolution & 
    \# Channels  &
    \# Layers\\
    $i$ &
    $\mathcal{\hat{F}}_i$ &
    $\mathcal{\hat{H}}_i\times\mathcal{\hat{W}}_i$ &
    $\mathcal{\hat{C}}_i$ &
    $\mathcal{\hat{L}}_i$ \\
    \midrule
    $1$ & 
    \multicolumn{1}{c}{MBConv6, k3x3} &
    \multicolumn{1}{c}{$28\times28$} &
    \multicolumn{1}{c}{80} &
    \multicolumn{1}{c}{3}  \\ 
    $2$ & 
    \multicolumn{1}{c}{MBConv6, k5x5} &
    \multicolumn{1}{c}{$14\times14$} &
    \multicolumn{1}{c}{112} &
    \multicolumn{1}{c}{3}  \\ 
    $3$ & 
    \multicolumn{1}{c}{MBConv6, k5x5} &
    \multicolumn{1}{c}{$14\times14$} &
    \multicolumn{1}{c}{192} &
    \multicolumn{1}{c}{4} \\
    $4$ & 
    \multicolumn{1}{c}{MBConv6, k3x3} &
    \multicolumn{1}{c}{$7\times7$} &
    \multicolumn{1}{c}{320} &
    \multicolumn{1}{c}{1} \\
    $5$ & 
    \multicolumn{1}{c}{Conv1x1 \& Pooling \& FC} &
    \multicolumn{1}{c}{$7\times7$} &
    \multicolumn{1}{c}{1280} &
    \multicolumn{1}{c}{1} \\
    \bottomrule
    \end{tabular}
    }
\caption{The eFUN-variant-B architecture. Each row describes a stage, $i$, with $\hat L_i$ layers, input resolution $\langle \hat H_i, \hat W_i \rangle$ and output channels $\hat C_i$.}
\label{tb:ablation_bottleneck}
\end{table}

In this section we conduct an ablation study with various alternative architectures to those presented above in order to validate our architecture design.
The first architecture considered is the standard EfficientNet-B0 architecture presented in \cite{tan2019efficientnet}, with a few key changes. First, the first convolution layer is removed and the number of strides is reduced accordingly, in order to keep the spatial dimension of the output layer the same as the original architecture. The modified architecture is presented in Table~\ref{tb:ablation_effnet_b0} and is denoted eFUN-variant-A. 
The second architecture was designed by following the modifications made to the ResNet-50 model in order to create the ResFUN architecture. In this architecture, the DCT inputs are fed directly to the stage in the original EfficientNet architecture which is designed to have inputs with the same spatial size as eFUN's DCT inputs ($28\times28$). The rest of the network is the same as the final stages of the standard EfficientNet-B0 architecture. The modified architecture is denoted eFUN-variant-B and presented in Table~\ref{tb:ablation_bottleneck}.

Our results, presented in Table~\ref{tb:ablation}, show that the original eFUN model outperforms both variants. We conclude that the balance between the width and resolution of the network is crucial and that a naive change to the architecture does not yield the same results. In particular, we find that EfficientNet-B0 is not suitable for DCT inputs due to the large difference in input resolution required. Thus, intelligent modifications are needed to design a smaller architecture able to exploit the strengths of the DCT representation.

\begin{table}
    \setlength{\tabcolsep}{2.2pt}
    \centering
    \resizebox{1.0\columnwidth}{!}{ 
    \begin{tabular}{c c c}
    \toprule
    Architecture & Top-1 Acc. & \# Parameters (M) \\
    \midrule
    eFUN &
    \multicolumn{1}{c}{77} &
    \multicolumn{1}{c}{4.2} \\
    eFUN-variant-A &
    \multicolumn{1}{c}{75} &
    \multicolumn{1}{c}{5.3} \\
    eFUN-variant-B &
    \multicolumn{1}{c}{74.9} &
    \multicolumn{1}{c}{5.6} \\
    \bottomrule
    \end{tabular}
    }
    \caption{Architecture ablation study, presented on the ImageNet dataset.}
\label{tb:ablation}
\end{table}

\subsection{Input Compression} \label{input compression section}

\begin{table}
    \setlength{\tabcolsep}{2.2pt}
    \centering
    \resizebox{1.0\columnwidth}{!}{ 
    \begin{tabular}{c c c c c}
    \toprule
    \# Input Channels & Y/Cb/Cr & Top-1 Acc. & \# Parameters (M) \\
    \midrule
    $192$ &
    \multicolumn{1}{c}{$64/64/64$} &
    \multicolumn{1}{c}{$77.01$} &
    \multicolumn{1}{c}{$4.23$}  \\ 
    $88$ & 
    \multicolumn{1}{c}{$64/12/12$} &
    \multicolumn{1}{c}{$76.02$} &
    \multicolumn{1}{c}{$3.99$} \\
    $64$ & 
    \multicolumn{1}{c}{$44/10/10$} &
    \multicolumn{1}{c}{$73.95$} &
    \multicolumn{1}{c}{$3.96$}   \\
    $48$ & 
    \multicolumn{1}{c}{$32/8/8$} &
    \multicolumn{1}{c}{$69.45$} &
    \multicolumn{1}{c}{$3.93$}   \\
    $24$ & 
    \multicolumn{1}{c}{$14/5/5$} &
    \multicolumn{1}{c}{$51.85$} &
    \multicolumn{1}{c}{$3.91$} \\
    \bottomrule
    \end{tabular}
    }
    \caption{The effects of dropping input channels on eFUN performance.}
\label{tb:performace_vs_num_channels}
\end{table}

In this experiment, we take a trained eFUN show the effects of discarding input channels on the model accuracy and size. To do so, during inference selected Y, Cb, and Cr channels are zeroed-out. 
We show that one can significantly reduce the number of frequency channels used to represent the input by keeping only the lowest frequency channels in each of Y, Cb, and Cr.

\begin{figure}
    \includegraphics[scale=0.4]{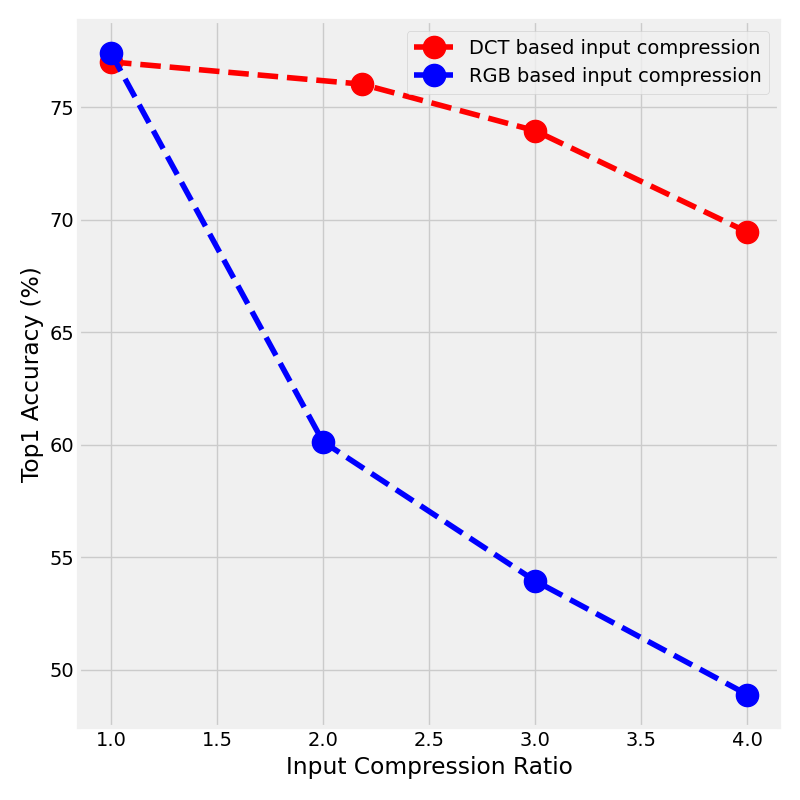}
    \caption{ImageNet accuracy vs. input compression ratio.}
    \label{imagenet_acc_vs_input_compression}
\end{figure}

Our results, presented in Table~\ref{tb:performace_vs_num_channels}, show that when using FUN, it is possible to significantly reduce the size of the input, while maintaining a relatively high accuracy. For example, dropping $54\%$ of the input channels (from $192$ to $88$) results in a $1\%$ drop in accuracy while dropping $75\%$ of the input channels results in a $7.5\%$ drop in accuracy. These results offer a desirable property: as no additional training is performed here, FUN models inherently support different possible operating points with regard to the input compression ratio. Figure~\ref{imagenet_acc_vs_input_compression} shows the effect of compression on the Top-1 accuracy for both our DCT eFUN model and an RGB EfficientNet-B0, where simple down-sampling is used to simulate compression for RGB-based models.

\begin{figure*}
\setlength{\tabcolsep}{1pt}
\centering
    \begin{tabular}{c c c}
        \includegraphics[width=0.3\textwidth]{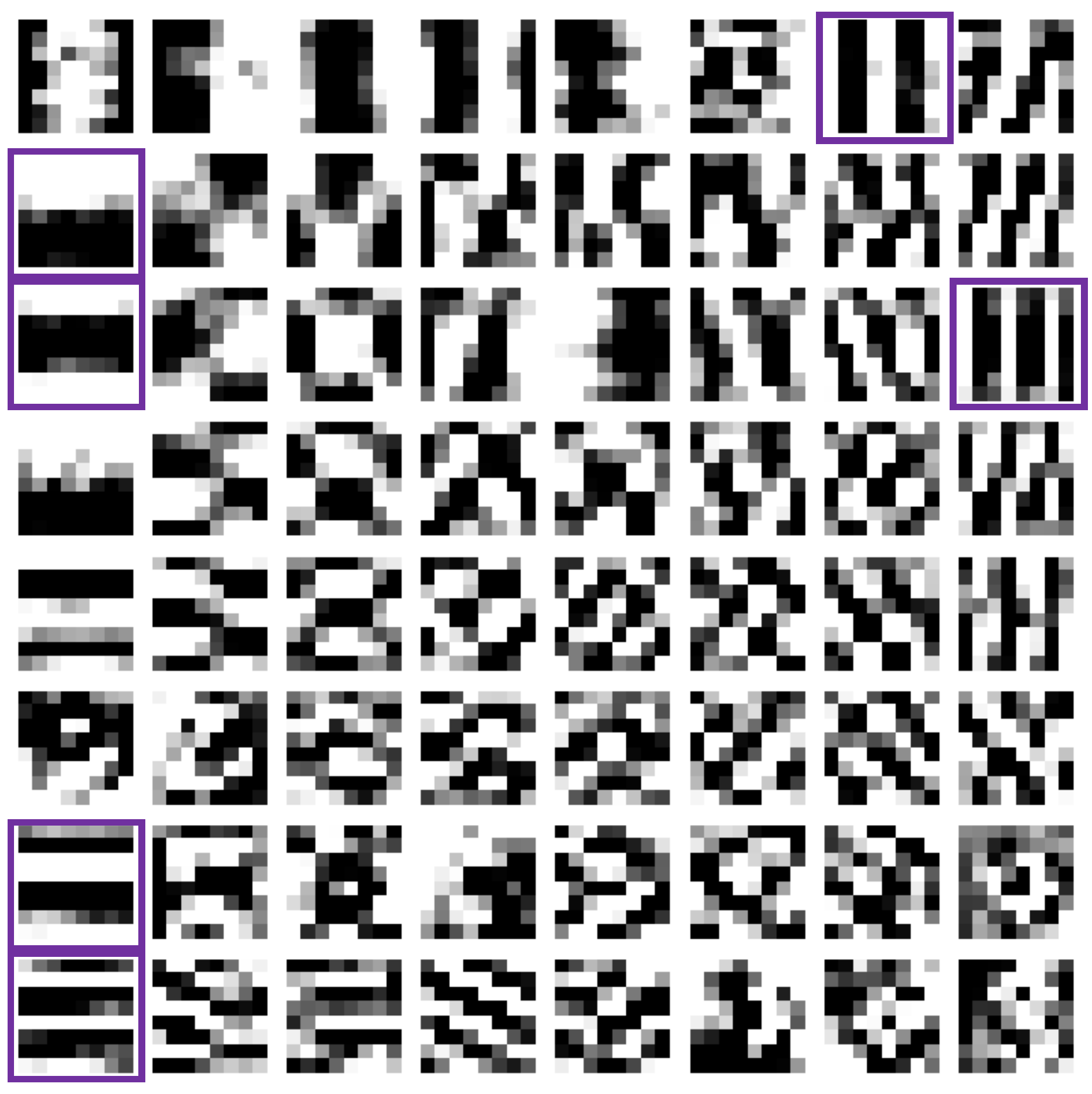}&
        \includegraphics[width=0.3\textwidth]{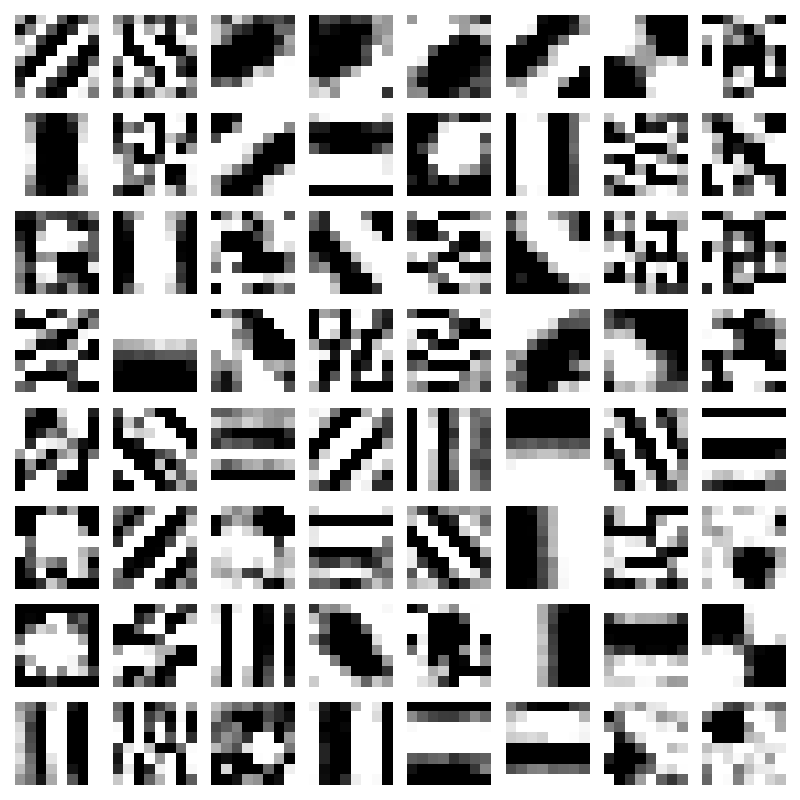}&
        \includegraphics[width=0.3\textwidth]{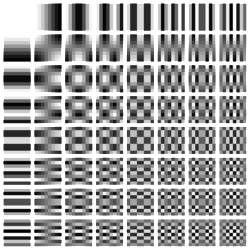}
    \end{tabular}
    \caption{Dct filters. From left to right: $\mbox{LeFUN}$, $\mbox{LeFUN}_{\mbox{e2e}}$, standard DCT.}
    \label{fig:dct_filers}
\end{figure*}

\subsection{A Learnable DCT Transformation}
The core advantage of FUN comes from the usage of the DCT representation, which is static and not learned.
In this experiment, we study whether the input DCT can be replaced with a learnable convolutional layer. In particular, the new convolutional layer receives as input a YCbCr image obtained from its RGB representation. It then transforms each $8\times8\times1$ input block into $1\times1\times64$ outputs. The final DCT output, of size $28\times28\times192$, is then fed into the original network, as before.

We propose two methods of learning this transformation:
\begin{enumerate}
    \item Adding a DCT-like convolutional layer before an already trained eFUN model and training only the added convolutional layer, this Learnable-eFUN variant is denoted LeFUN.
    \item Adding a DCT-like convolutional layer before the eFUN architecture and training the entire network end-to-end, denoted $\mbox{LeFUN}_{\mbox{e2e}}$.
\end{enumerate} 
Intuitively, the second approach implements an "encoder-decoder" architecture, where the decoder is a fixed eFUN network and the encoder is trained to produce inputs similar to those expected by the decoder.

The results achieved by each of the methods are presented in Table~\ref{tb:res_learnable_dct}. The results show that $\mbox{LeFUN}_{\mbox{e2e}}$ provides the best accuracy, while resulting in a slightly larger model compared to the standard eFUN model, and $\mbox{LeFUN}$ provides the worst accuracy by a large margin.
The filters learned by each of the three methods are presented in Figure~\ref{fig:dct_filers}.
One can see that some of the filters learned by our new layers seem similar to those of the standard DCT, as they include vertical and horizontal lines in some frequency.
However, most filters still look rather different than those of the DCT, showing the usefulness of using the static DCT filters, which allows an higher level of understanding and has meaningful ordering of the filters.

Our results show that using a DCT-like learned convolutional layer for RGB inputs can provide results similar to those of eFUN models in terms of model size, latency, and top-1 accuracy. However, using a learned transformation makes it difficult to analyze the importance of each input channel and does not provide inherent support for dynamic input compression presented in Section~\ref{input compression section}.

\begin{table}
    \setlength{\tabcolsep}{2.2pt}
    \centering
    \resizebox{1.0\columnwidth}{!}{ 
    \begin{tabular}{l c c}
    \toprule
    Model & Top-1 Acc.& \# Parameters (M)\\
    \midrule
    eFUN &
    \multicolumn{1}{c}{77} &
    \multicolumn{1}{c}{4.2} \\ 
    LeFUN  &
    \multicolumn{1}{c}{72.8} &
    \multicolumn{1}{c}{4.3} \\ 
    $\mbox{LeFUN}_{\mbox{e2e}}$ &
    \multicolumn{1}{c}{77.2} &
    \multicolumn{1}{c}{4.3} \\ 
    \bottomrule
    \end{tabular}
    }
    \caption{Comparison of fixed and learnable transforms.}
\label{tb:res_learnable_dct}
\end{table}

\section{Conclusions} \label{Conclusions}
In this paper we have presented the FUN family. These architectures were designed, using modern building blocks, to utilize DCT inputs instead of the standard RGB ones. We show that using this representation one can create architectures that are much more efficient, in terms of size and latency, while still maintaining the accuracy of RGB-based models. The effectiveness of this representation is then shown using an extensive set of experiments. We hope that, apart from its direct impact, FUN would inspire more research into the effectiveness of the frequency-domain for training neural networks.

{\small
\bibliographystyle{ieee_fullname}
\bibliography{egbib}
}

\end{document}